# Non-parametric convolution based image-segmentation of ill-posed objects applying context window approach

Upendra Kumar[1], Tapobrata Lahiri[2*] and Manoj Kumar Pal[2]

[1]Gautam Buddh Technical University, Lucknow, India-212211
[2]Indian Institute of Information Technology, Allahabad, India- 211012
Email: tlahiri@iiita.ac.in

**Abstract:**
Context-dependence in human cognition process is a well-established fact. Following this, we introduced the image segmentation method that can use context to classify a pixel on the basis of its membership to a particular object-class of the concerned image. In the broad methodological steps, each pixel was defined by its context window (CW) surrounding it the size of which was fixed heuristically. CW texture defined by the intensities of its pixels was convoluted with weights optimized through a non-parametric function supported by a backpropagation network. Result of convolution was used to classify them. The training data points (i.e., pixels) were carefully chosen to include all variety of contexts of types, i) points within the object, ii) points near the edge but inside the objects, iii) points at the border of the objects, iv) points near the edge but outside the objects, v) points near or at the edge of the image frame. Moreover the training data points were selected from all the images within image-dataset. CW texture information for 1000 pixels from face area and background area of images were captured, out of which 700 CWs were used as training input data, and remaining 300 for testing. Our work gives the first time foundation of quantitative enumeration of efficiency of image-segmentation which is extendable to segment out more than 2 objects within an image.

**Introduction:**
It is a common and existing knowledge that to recognize an object of interest, we should first segment out the object from the background to reject noisy or redundant information on the object from the decision making process on it. The protocol is generally referred as face-detection in the paradigm of face as our object of interest. In this direction we found the existing techniques are not efficient because of their lack of tolerance to change in diverse backgrounds. Taking cue from this problem we have designed and applied a simple context window description of texture for each pixel to fix their membership to either object, i.e., face or background, i.e., diverse background properties excluding existence of any other face.

Segmentation is an important process utilized in many image processing applications. Partitioning of a given image into different regions of interest is defined as segmentation. These regions are set of pixels. The number of region of interest depends on the need of any application. The purpose of segmenting out any object image is to simplifying the image representation into something more meaningful, which can be analyzed in better and easier way. In image processing application, the efficiency and accuracy depends on the algorithm or methodology used for the classification. Many of these techniques suffer from incorrect classification due to not supporting features chosen. Filtering method and morphological approach used for image enhancement and restoration gave incorrect classification results because of incorrect representation of the extracted feature or information content [**6**]. The severity of this problem gets increased in the noisy environment, because of incorrect classification using noisy feature. Neural network has been proven a good classifier for the data with noisy content provided that feature of object has been extracted correctly [**7-8**]. Many image segmentation work has been done using either pixel-based or context window-based classification. The approach using pixel-based classification is very time taking while the efficiency of the approach using context window based classification depends on the accuracy of the context window chosen. This approach has been used in document image segmentation [**9**]. Spatial context concept gives more important information for better segmentation of object of interest where object is represented by set of pixels [**10**]. It has been observed that the apparent brightness of any object is not completely based on its luminance but also based on the context in which it is embedded **[11-12].** The inclusion of contextual information in recognizing any object gives better recognizing efficiency. Many sources of context information have been discussed and proposed by Divvala et al. **[14]**. One of the important sources of context is local pixel context, which in this study is termed as Context Window based Texture of Pixel (CWTP). This encapsulates the basic conception around the unit pixel of any image that carries useful information which helps out in getting better recognition ability. A learning based approach has been shown and applied in segmenting out the block of postal which contains the address of recipient **[15].**

---

[*] Corresponding author

**Methodology:**

**Collection of face image:**

Face images of different persons were collected by digital camera (Model (Canon –Model: Power Shot S50). Database comprises of 10 color images for 10 different subjects.

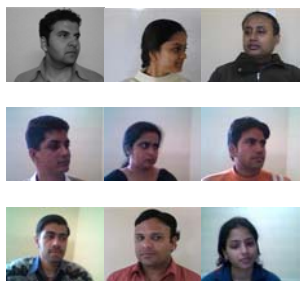

Figure 1: Sample frames in RGB of different person from database

**Preprocessing:**

For enhancement of the quality of image and to make it more suitable for further processing so that effective results can be obtained, we performed following preprocessing on frames (Gonzales & Woods, 2002) [**16**]

  i) Conversion from Color to Gray scale.

ii) Application of context window based concept using backpropagation network by inputting context window based texture of pixels (CWTP) as a feature for segmentation of region of interest [**17**].

**Face detection by segmentation of face images by applying context window based texture of pixels (CWTP) as feature to backpropagation network (BPN)**

Each image was converted to gray scale for further processing. Part of images representing faces were segmented out from background using an artificial neural network based texture-segmentation protocol. In this protocol each pixel was defined by its context window (CW) surrounding it the size of which was fixed as 9x9. The following figure 1 shows how to get the pixel data from both class (face part and background).

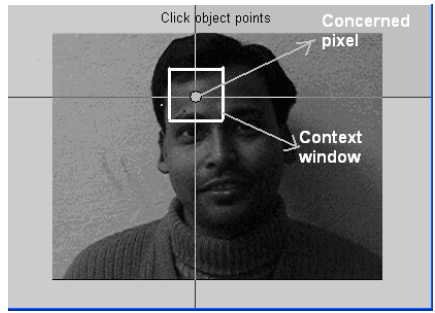

Figure 2: Selection of pixel for both face and background image segmentation where a typical CW for the concerned pixel has been shown.

CW texture defined by the intensities of its pixels was fed as input to the ANN to yield a (1, -1) output for object, (-1, 1) output for background. Finally the object and background pixel values were assigned as 1 and 0 respectively. The training data points (i.e., pixels) were carefully chosen to include all variety of contexts of types, i) points within the object, ii) points near the edge but inside the objects, iii) points at the border of the objects, iv) points near the edge but outside the objects, v) points near or at the edge of the image frame. Moreover the training data points were selected from randomly picked different image types as described in image acquisition section. CW texture information for 1000 pixels from face area and background area of images were captured, out of which 700 CWs were used as training input data, and remaining 300 for testing. The ANN we used for this purpose was a four layered feed-forward backpropagation network having 81, 18, 10, and 2 neurons in input, first hidden, second hidden, and output layer respectively.

**Mathematical formalism:**

In the mathematical formalism we may consider pixel-intensities within a context window of size 9X9 to represent the membership of a particular pixel as

$\{x_{i,j}\}_{i,j=1}^{9}$ .This was eventually considered to a single dimensional array as:

$$\{x_{i,j}\}_{i,j=1}^{9} \rightarrow \{y_k\}_{k=1}^{81}$$

$Y_k$'s were served as input to BPN to yield output O = $\{O_i\}_{i=1}^{2}$

Where as:

If , $O_1 > O_2$ implies face pixel.

And $O_2 >= O_1$ implies background pixel.

**BPN Architecture used for classification:**

Classification was done using Backpropagation algorithm. The BPN architecture consists of two hidden layers apart from input and output layers. While the input layer comprises of 81 inputs, the first and second hidden layers consist 18 and 10 neurons respectively. The last i.e. the output layer contains two neurons to represent two classes of our interest (as in Figure 3). The conditions and parameters for ANN classification were kept as follows: Network type: Feed Forward backpropagation, Transfer function: TANSIG, Training Function: TRAINLM, Performance function: MSE (mean square error) where, MSE = mean (error^2), where error = desired output – actual output.

After training we obtained the optimized values of three weight matrices $W_1$ of size 18 × 81, $W_2$ of size 10 × 18 and $W_3$ of size 2× 10, and three bias matrices $b_1$ of size 1 × 18, $b_2$ of size 1 × 10 and $b_3$ of size 1 × 2.

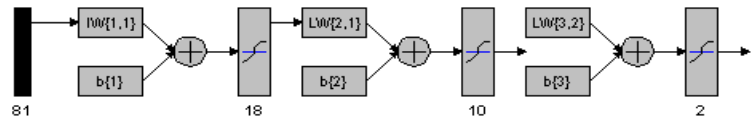

Figure 3: BPN Architecture used for Image segmentation

**Result:**

**Segmentation result by using Gray scale image:**

Result of segmentation has been cited in the following figure 2 and table 1.

| (a) original face image with | (b) its segmented form in | (c) Its segmented form in gray |
|---|---|---|

| background | binary | scale |
|---|---|---|

Figure 4: Result of segmentation of Image using 9X9 size context window (a) original face image with background and (b) its segmented form in binary. (c) Its segmented form in gray scale.

**Result by using ANN with Context window size 9X9:**

Number of neurons in each layer were taken 81, 18,10,2 respectively.

|  | Training |  | Testing |  |
|---|---|---|---|---|
| Total number of pixels | 700 | Classification efficiency is  78.43 | 300 | Classification efficiency is  80.33 |
| Total number of pixels correctly classified | 549 |  | 241 |  |

Table 1: Efficiency in pixel-wise segmentation by CWTP fed to BPN

**Result by using ANN with Context window size 7X7:**

Number of neurons in each layer were taken 49, 25,10,2 respectively.

|  | Training |  | Testing |  |
|---|---|---|---|---|
| Total number of pixels | 700 | Classification efficiency is  74.29 | 300 | Classification efficiency is  70.67 |
| Total number of pixels correctly classified | 520 |  | 212 |  |

Table 2: Efficiency in pixel-wise segmentation by CWTP fed to BPN

**Result by using ANN with Context window size 11X11:**

Number of neurons in each layer were taken 49, 25,10,2 respectively.

|  | Training |  | Testing |  |
|---|---|---|---|---|
| Total number of pixels | 700 | Classification efficiency is | 300 | Classification efficiency is |

| Total number of pixels correctly classified | 530 | 75.71 | 209 | 69.67 |

Table 3: Efficiency in pixel-wise segmentation by CWTP fed to BPN

Result of segmentation using Gabor filter method [5] has been cited in the following figure 3.

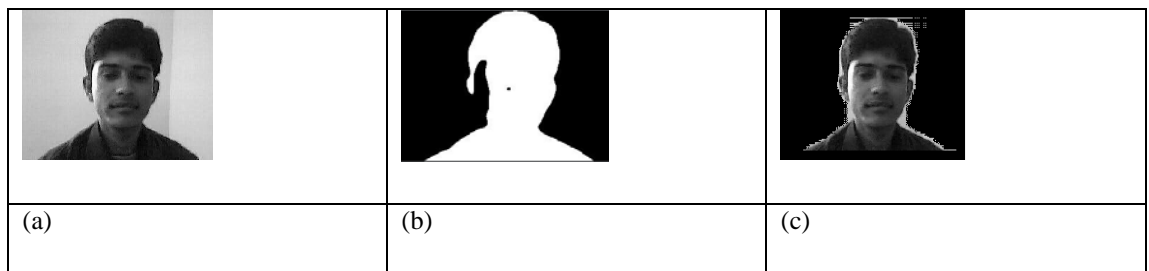

| (a) | (b) | (c) |

Figure 5: Result of segmentation using Gabor filter (a) original face image with background and (b) its segmented form in binary.

**Segmentation result by using RGB image:**

**Result by using ANN with Context window size 5X5:**

Result of segmentation has been cited in the following figure 6 and table 4. Number of neurons in each layer were taken 81, 18,10,2 respectively.

|  | Training |  | Testing |  |
|---|---|---|---|---|
| Total number of pixels | 700 | Classification efficiency is 99.86 | 300 | Classification efficiency is 87.00 |
| Total number of pixels correctly classified | 699 |  | 261 |  |

Table 4: Efficiency in pixel-wise segmentation by CWTP fed to BPN

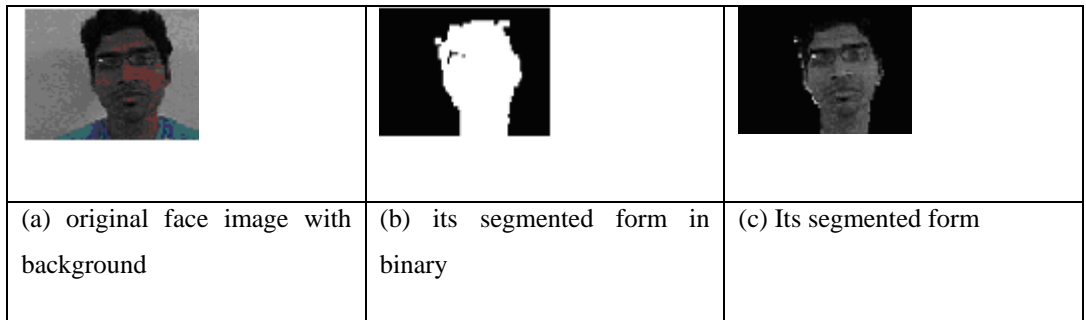

| (a) original face image with background | (b) its segmented form in binary | (c) Its segmented form |

Figure 6: Result of segmentation of Image using 5X5 size context window (a) original face image with background and (b) its segmented form in binary. (c) Its segmented form.

**Result by using Nearest Neighbor Concept with Context window size 5X5:**

Result of segmentation has been cited in the following figure 6 and table 4. Number of neurons in each layer were taken 81, 18,10,2 respectively.

|  | Training |  | Testing |  |
|---|---|---|---|---|
| Total number of pixels | 700 | Classification efficiency is  78.57 | 300 | Classification efficiency is  75.00 |
| Total number of pixels correctly classified | 550 |  | 225 |  |

Table 4: Efficiency in pixel-wise segmentation by CWTP fed to BPN

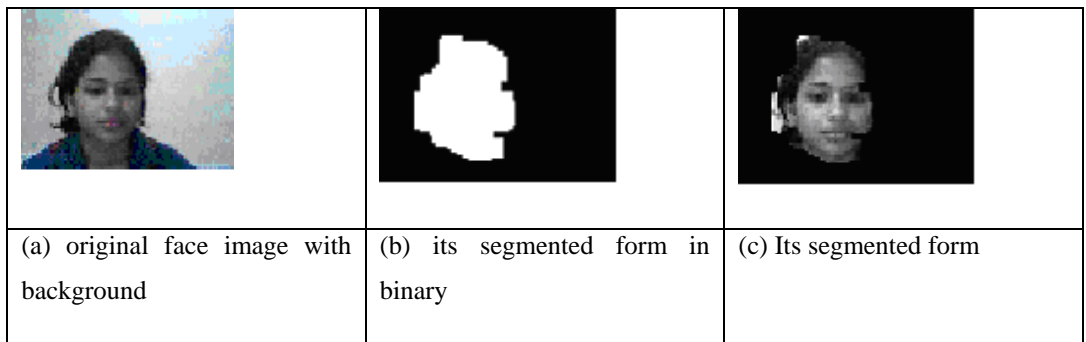

| (a) original face image with background | (b) its segmented form in binary | (c) Its segmented form |

Figure 7: Result of segmentation of Image using 5X5 size context window (a) original face image with background and (b) its segmented form in binary. (c) Its segmented form.

**Discussion:**

The protocol described above was found to work much better than the existing methods, e.g., Gabor filter based method and other methods (result was not shown) [5]. However being an unsupervised filter, Gabor filter can even segment out more than 2 objects (i.e., face and background) for which drawing binary definition of the objects as background or face will be difficult. Therefore we have applied a supervised texture detection technique which is similar to a convolution filter having optimized ANN weights as its coefficients. However, in case of nearest neighbor approach minimally distant neighboring context pixel defines the pixel property. The context window concept was however borrowed from the work of Qian et al [2] where the window was a single dimensional array. In our case however we have modified the method to use 2 dimensional arrays as context after converting an array of size 9x9 to single dimensional array of size 1x81 only. The simple technique found to be powerful enough to capture the membership of a pixel for object, i.e., face or background as shown in figure 1,2,3 and 4 and table 1,2,3 and 4. Also the accuracy of the segmentation model appeared to be dependent on the choice of types of pixels representing data for various possible cases. A mere selection of data inside and outside of the object (i.e., considering it as a two case problem) was not found to be sufficient and therefore we have considered many more different cases (as described in the methodology section). The specific importance of context is that its size should be least and even then it should cover overall attribute of a pixel. The context is beneficial mainly on two counts, first to capture texture and second to capture color and grey-ness within the local surroundings of a pixel. Therefore, in the preprocessing part of choosing training pixels (i.e., training contexts) care should be taken to cover all the contexts, e.g., confusing facial boundary of both internal (e.g., boundary between forehead and hair) and external types (e.g., boundary between chick with external background) variety of pixels. Since these contexts are used for training of pixel-classifiers, missing any one of them may be proved to be costly. In this study, the efficiency profile of different classifier shows that quantitative efficiency is quite correlated to visual representation of segmentation-results. This is quite evident in case of use of nearest neighbor classifier. The reason for observed little discrepancy for backpropagation classifier may be due to the over-fitting problem.

**Conclusion:**

A context window based texture representation of pixels was used to define membership of the pixels to face and background. The context window concept also appeared to follow human cognition module where a human detects different textures on the basis of their context area only [3]. Therefore, the one more benefit of approach of this work is found to be its ability to give quantitative measure of efficiency for the first time in the paradigm of image segmentation. The other benefit in the paradigm of face recognition is that most of the face recognition method utilizes box-type image-frame covering face as data that includes background information also within this box. The work described in this paper-work indicates that the contribution of background noise can be further minimized that logically appears to help in improving efficiency of face recognition.